\documentclass[english]{cccconf}
\usepackage[comma,numbers,square,sort&compress]{natbib}
\usepackage{epstopdf}
\usepackage{hyperref}
\usepackage{subfigure}
\usepackage{amssymb}
\usepackage{amsmath}
\usepackage{color}
\hypersetup{hidelinks,
	colorlinks=true,
	allcolors=blue,
	pdfstartview=Fit,
	breaklinks=true}
\begin{document}

\title{EdgeYOLO: An Edge-Real-Time Object Detector}

\author{Shihan Liu\aref{bitauto,bitcqic},
        Junlin Zha\aref{bitcqic},
        Jian Sun\aref{bitauto,bitcqic}, 
        Zhuo Li\aref{bitauto,bitcqic}, 
        and
        Gang Wang\aref{bitauto,bitcqic}}


\affiliation[bitauto]{National Key Lab of Autonomous Intelligent Unmanned Systems, Beijing Institute of Technology, Beijing 100081, China         \email{liushihan@bit.edu.cn; sunjian@bit.edu.cn;  zhuoli@bit.edu.cn; gangwang@bit.edu.cn}}


\affiliation[bitcqic]{Beijing Institute of Technology Chongqing Innovation Center, Chongqing 401135, China
        \email{jlzha8101@163.com}}

\maketitle

\begin{abstract}
This paper proposes an efficient, low-complexity and anchor-free object detector based on the state-of-the-art YOLO framework, which can be implemented  in real time on edge computing platforms. We develop an enhanced data augmentation method to effectively suppress overfitting during training, and design a hybrid random loss function to improve the detection accuracy of small objects. Inspired by FCOS, a lighter and more efficient decoupled head is proposed, and its inference speed can be improved with little loss of precision. Our baseline model can reach the accuracy of \textbf{50.6$\%$ AP$_{50:95}$} and \textbf{69.8$\%$ AP$_{50}$} in MS COCO2017 dataset, \textbf{26.4$\%$ AP$_{50:95}$} and \textbf{44.8$\%$ AP$_{50}$} in VisDrone2019-DET dataset, and it meets real-time requirements (FPS$\geq$30) on edge-computing device Nvidia Jetson AGX Xavier. And as is shown in Fig.\ref{fig0}, we also designed lighter models with less parameters for edge computing devices with lower computing power, which also show better performances. Our source code, hyper-parameters and model weights are all available at \textcolor{blue}{https://github.com/LSH9832/edgeyolo}.
\end{abstract}


\footnotetext{The work was supported in part by the National Natural Science Foundation of China under Grants 61925303, 62173034,  62088101. }

\section{Introduction}

As computing hardware performance continuously improves, computer vision technology based on deep neural networks has ushered rapidly in the last ten years, where object detection consists of an important element for applications in autonomous intelligent systems \cite{ais}. Nowadays, there are two mainstream object detection strategies. One is a two-stage strategy represented by the R-CNN series \cite{rcnn, fasterrcnn}, and the other one is a one-stage strategy with YOLO \cite{yolo1, yolo2, yolo3} as one of the most popular frameworks. For two-stage strategies, a heuristic method or regional suggestion generation method is used to obtain multiple candidate boxes in the first stage, and then these candidate boxes are screened, classified and regressed in the second stage. One-stage strategies give results in an end-to-end manner, where the object detection problem is transformed into a global regression problem. Global regression is not only capable of simultaneous assignment of place and category to multiple candidate boxes, but also of enabling models to get a clearer separation between object and background.

Models using a two-stage strategy perform a little better compared to those with a one-stage strategy on common object detection datasets such as MS COCO2017 \cite{coco}. Nevertheless, due to the inner limitations of the two-stage framework, it is far from meeting the real-time requirements on conventional computing devices, and it might face the same situation on most high-performance computing platforms. In contrast, one-stage object detectors can keep a balance between real-time indicators and performance. Thus, they are more concerned by researchers, and the YOLO series algorithm is updated iteratively at a high speed. Updates from YOLOv1 to YOLOv3 \cite{yolo1, yolo2, yolo3} are mainly improvements to the underlying framework structure, and most of the later mainstream versions of YOLO focus on improving precision and inference speed. Moreover, their optimization test platforms are mainly large workstations with high-performance GPUs. However, their state-of-the-art models usually run in unsatisfactorily low FPS on these edge computing devices. For this reason, some researchers proposed network structures with less parameters and lighter structures, such as MobileNet and ShuffleNet, to replace the original backbone networks, such that better real-time performance can be achieved on mobile devices and edge devices at the expense of some precision. In this paper, we aim to design an object detector that has decent precision and can run on edge devices in real time.

\begin{figure}
	\centering
	\includegraphics[width=\linewidth]{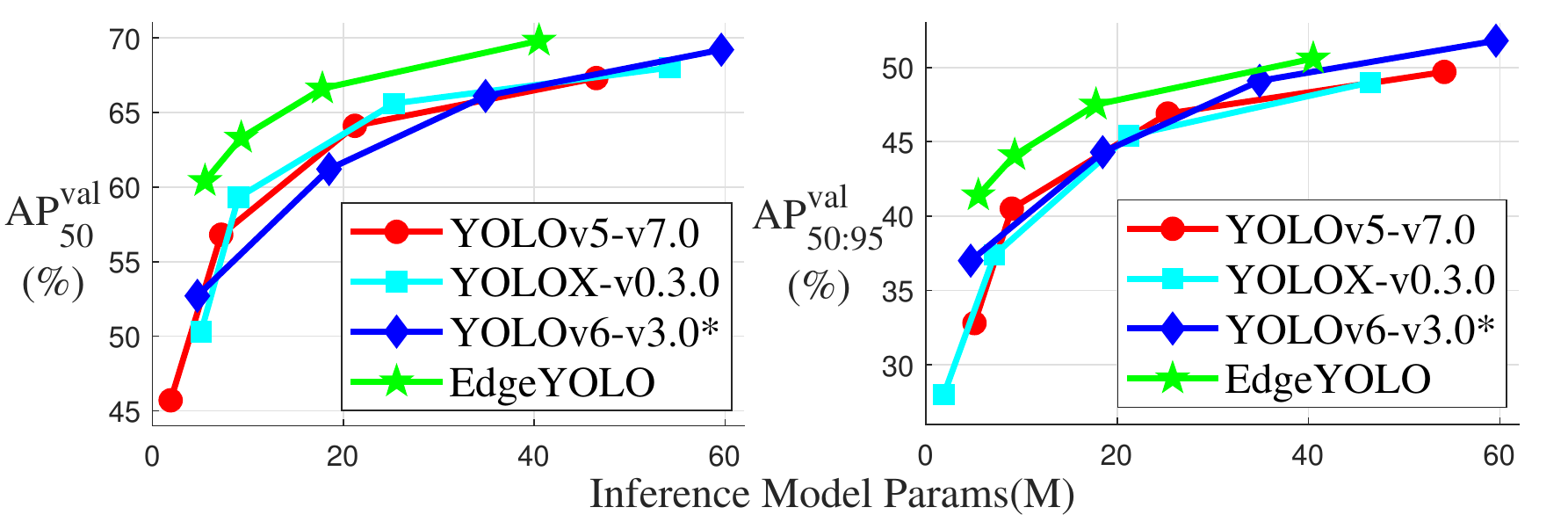}
	\caption{Comparison with some popular real-time object detectors. For fair comparison, \textbf{*} represents the results without using self-distillation\cite{yolo6_3} trick during training.}
	\label{fig0}
\end{figure}

The contributions of this paper are summarized as follows: i) An anchor-free object detector is designed, which can run on edge devices in real time with an accuracy of 50.6$\%$ AP in  MS COCO2017 dataset; ii) A more powerful data augmentation method is proposed, which further ensures the quantity and validity of training data; iii) Structures that can be re-parameterized are used in our model to reduce inference time; and, iv) A loss function is designed to improve the precision on small objects.

\begin{figure*}
  \centering
  \subfigure[Regular Mosaic+Mixup results]{
		\begin{minipage}[t]{1.0\linewidth}
			\centering
			\includegraphics[width=\linewidth]{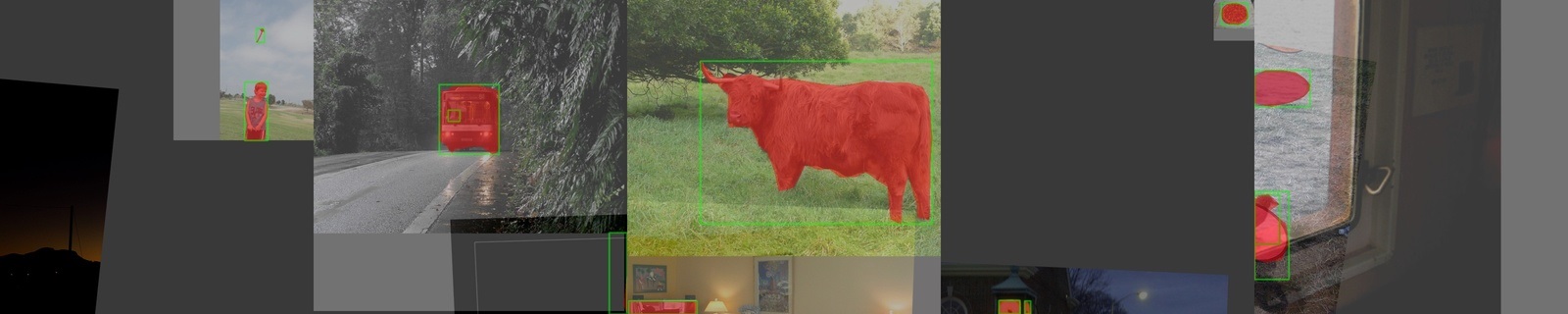}
		\end{minipage}
	}%

	\subfigure[Our data augmentation results]{
		\begin{minipage}[t]{1.0\linewidth}
			\centering
			\includegraphics[width=\linewidth]{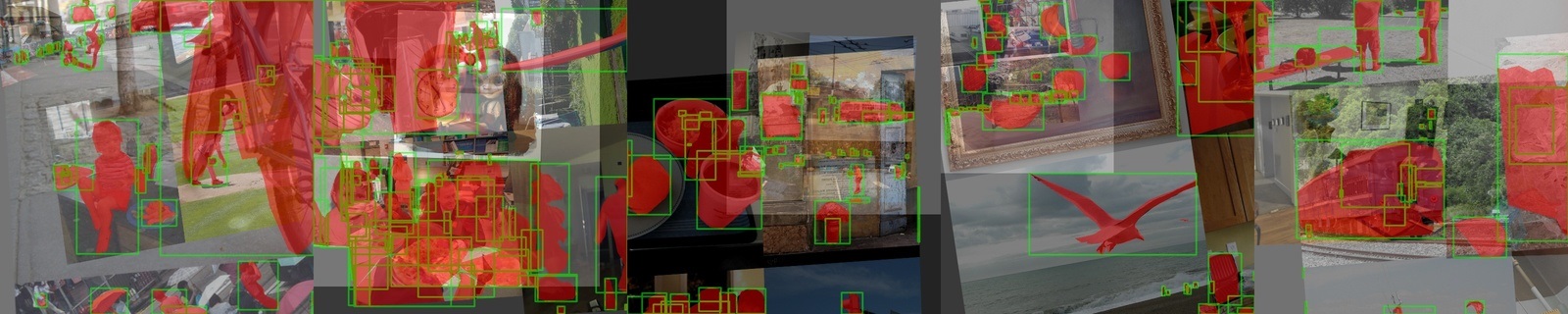}
		\end{minipage}
	}%
	\centering
	\caption{Random data augmentation inevitably causes some labels to be invalid, such as the lower right corner of the second figure and the lower left corner of the third picture in (a). Although there are boxes, they cannot give effective object information. A too small number of labels has an apparent negative impact on training, which can be avoided by increasing the number of effective boxes as in (b).}
	\label{fig1}
\end{figure*}

\section{Related Work}

\subsection{Anchor-free Object Detector}

Since the advent of YOLOv1, the YOLO series have been leading the field of real-time object detection for a long time. There are some other excellent detectors, such as SSD \cite{ssd}, FCOS \cite{fcos}, etc. When testing FPS in an object detection task, most previous studies only calculated the time cost of model inference, while a complete object detection task contains three parts: pre-process, model inference and post-process. Since pre-processing can be completed during video streaming, the post-processing time cost should be included when calculating the FPS of object detection. On a high-performance GPU workstation or server, pre-process and post-process only take up a small proportion of the time, whereas it takes even more than ten times the latency on an edge computing device. Thus, reducing post-processing computation can get a significant speed-up for edge computing devices. When using an anchor-based strategy, time latency in post-processing is almost proportional to the number of anchors of each grid cell. Anchor-based YOLO series usually allocates 3 anchors to each grid cell. Compared with those anchor-based frameworks, an anchor-free detector can save more than half of the time in the post-processing part.

To ensure the real-time performance of the detector on edge computing devices, we choose to build an object detector based on anchor-free strategy. Currently there are two main types of anchor-free detectors, one of which is anchor-point-based and the other is keypoint-based. In this paper we adopt an anchor-point-based paradigm.

\begin{figure*}
  \centering
  \subfigure[]{
		\begin{minipage}[t]{0.27\linewidth}
			\centering
			\includegraphics[width=0.65\linewidth]{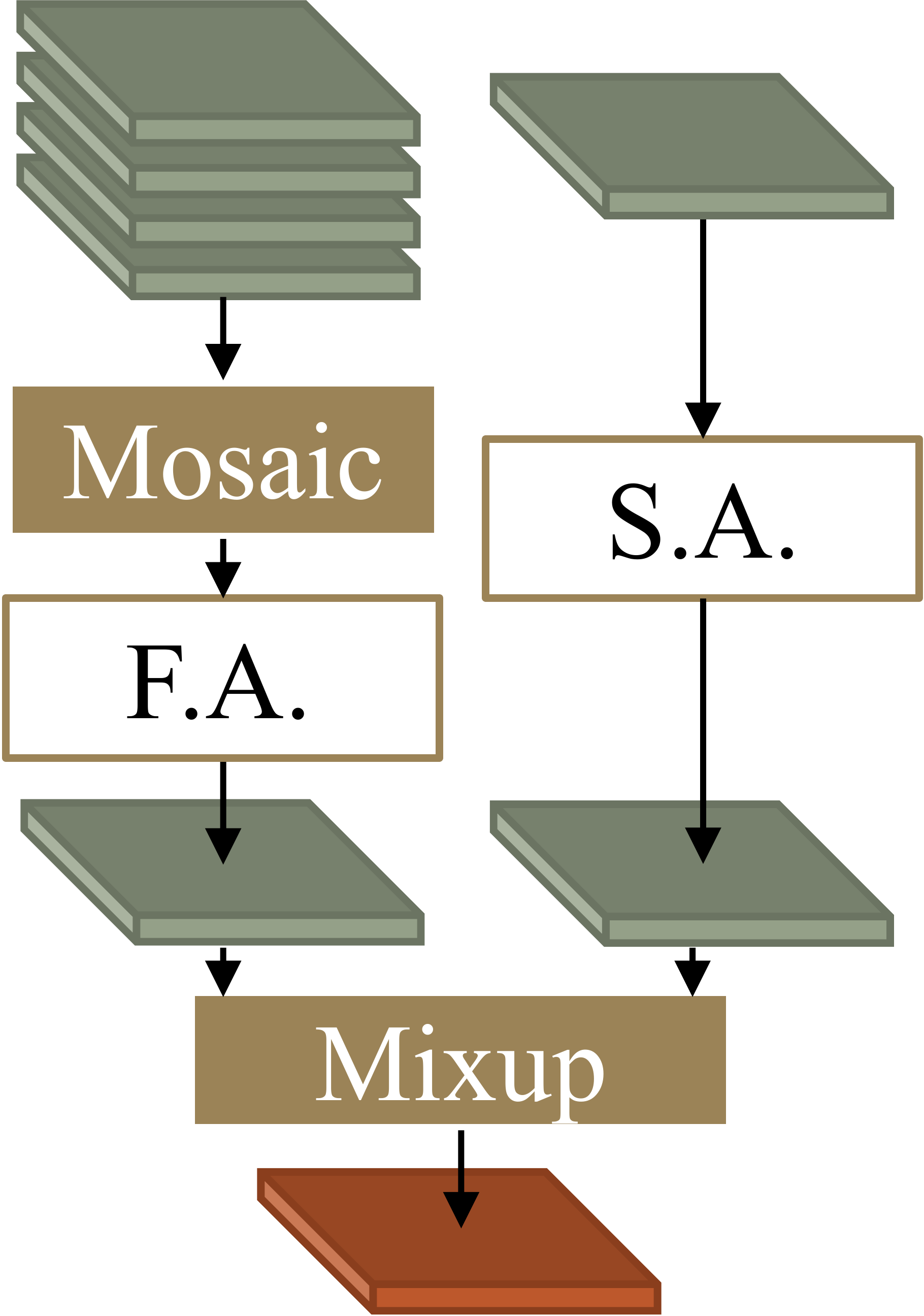}
		\end{minipage}
	}%
	\subfigure[]{
		\begin{minipage}[t]{0.27\linewidth}
			\centering
			\includegraphics[width=0.65\linewidth]{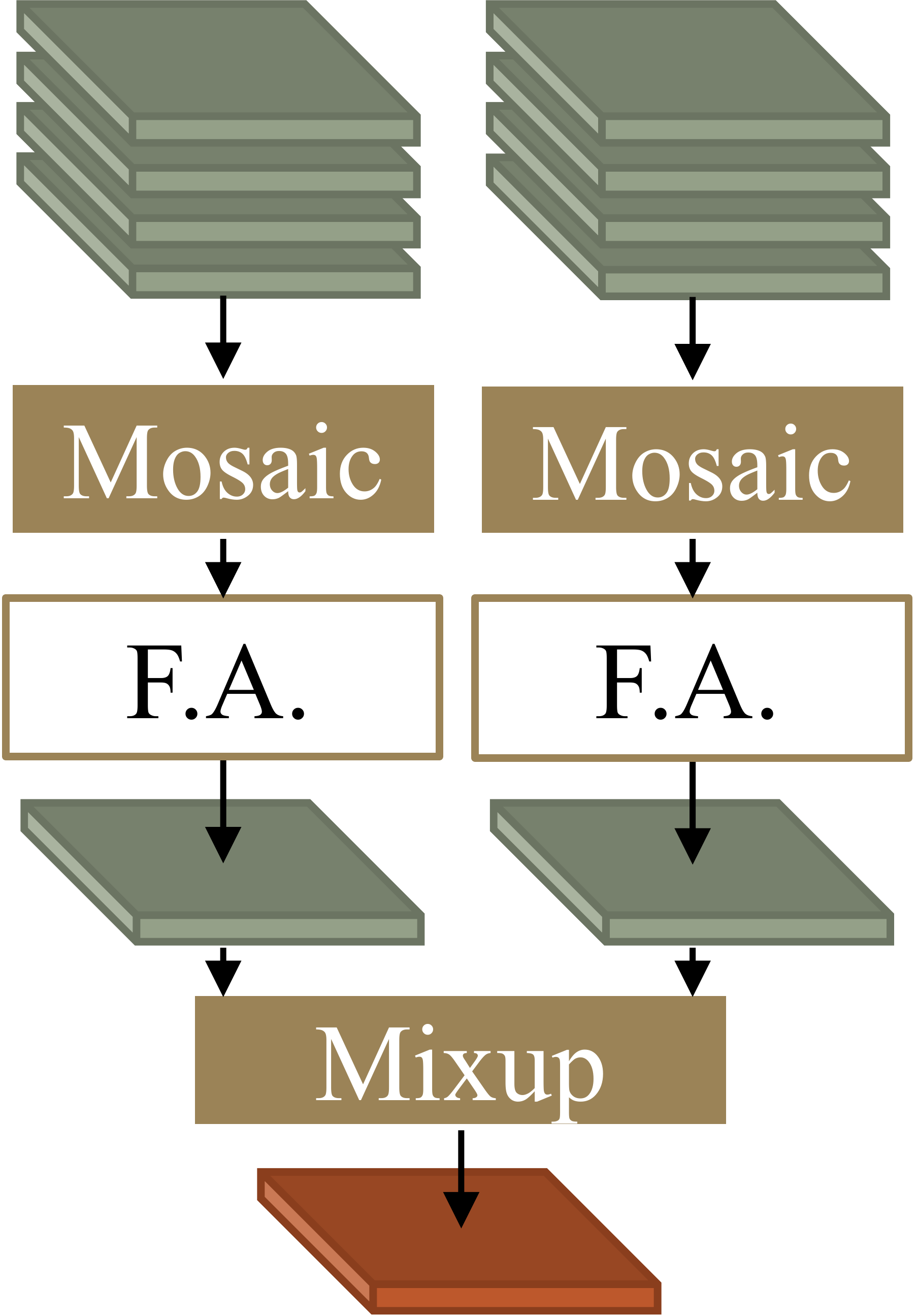}
		\end{minipage}
	}%
	\subfigure[]{
		\begin{minipage}[t]{0.46\linewidth}
			\centering
			\includegraphics[width=0.683\linewidth]{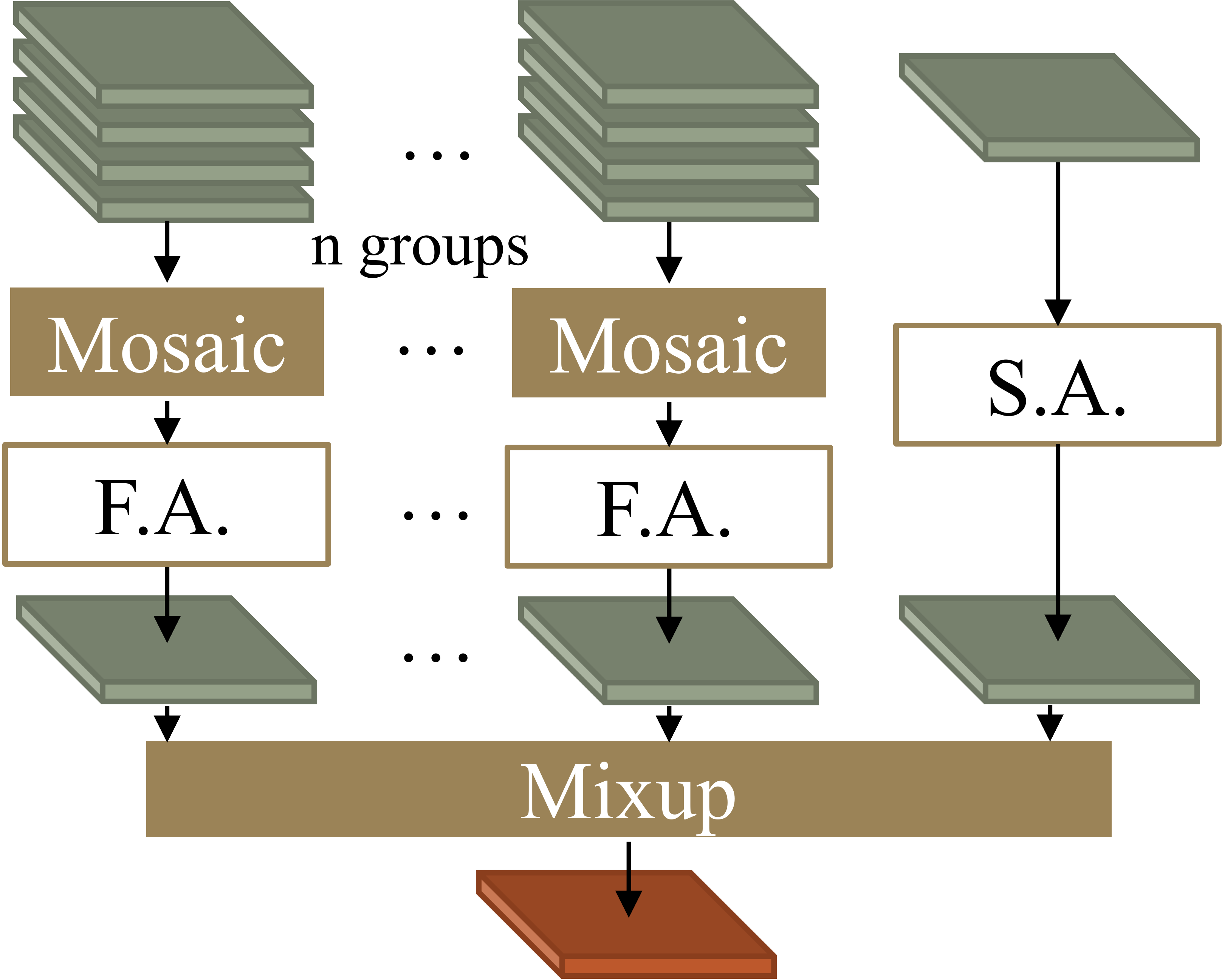}
		\end{minipage}
	}%
	\centering
	\caption{Different solutions for data augmentation (F.A.: Full Augmentation, S.A.: Simple Augmentation without large scale transformation). Shown in (a) \cite{yolox} and (b) \cite{yolo7}, using a fixed number of images in data augmentation is not suitable for all kinds of datasets. By using the method shown in (c), we can give flexible solutions for over-fitting problems.}
	\label{fig2}
\end{figure*}

\balance
\subsection{Data Augmentation}

Data augmentation is an essential data processing step in the training of neural networks. Rational use of data augmentation methods can effectively alleviate over-fitting of models. For image datasets, geometric augmentation (random cropping, rotation, mirroring, scaling, etc) and photometric augmentation (HSV $\&$ brightness adjustment) are often applied to a single image. These basic augmentation methods are often used before or after multi-image mixing and splicing. At present, mainstream data augment technologies, such as Mosaic \cite{yolo4}, Mixup \cite{mixup}, CopyPaste \cite{copypaste}, etc., put the pixel information of multiple pictures in the same picture through different methods to enrich the image information and reduce the probability of overfitting.

As is shown in Fig.\ref{fig1}(b), we design a more flexible and powerful combined-augmentation method, which further ensures the richness and validity of the input data.

\subsection{Model Reduction}

With model reduction, the computing cost is reduced, which can effectively improve the model inference speed. Model reduction methods can be divided into two categories: lossy reduction and lossless reduction. Lossy reduction usually builds smaller networks by reducing the number of network layers and channels. Lossless reduction integrates and couples multiple branch modules to build a more streamlined equivalent module by re-parameterizing techniques \cite{repvgg}. Lossy reduction achieves faster speed by sacrificing accuracy, and since coupled structure tends to reduce training effectiveness, the re-parameterization method is generally used for inference after model training is completed. 

By combining the lossy and lossless reduction methods, this paper builds several models of different sizes (shown in Fig.\ref{fig0}) to fit edge devices with different computing power, and speed up the model-inference process.

\subsection{Decoupled Regression}

From YOLOv1 to YOLOv5 \cite{yolo1, yolo2, yolo3, yolo4, yolo5}, for each feature map with a different scale, the regression for obtaining the location, category and confidence of objects uses a unified set of convolution kernels. In general, different tasks use the same convolution kernel if they are closely related. However, relations between the object's location, confidence and category are not close enough in numerical logic. Moreover, relevant experiments have proved that, compared with the direct unified regression detection head for all tasks, using a decoupled regression detection head \cite{fcos, yolox} can achieve a better result, and accelerate the loss convergence. Nonetheless, a decoupled head brings extra inference costs. As an improvement, we design a lighter decoupled head with joint consideration of inference speed and precision of the model.

\subsection{Small Object Detecting Optimization}

The problem of small object detection has been widely concerned since the beginning of object detection research. As the proportion of an object in the image decreases, the pixel information used to express the object decreases. A large object often occupies dozens or even hundreds of times of information compared to a small one, and the detection precision of small objects is often significantly lower than that of large objects. Moreover, this gap cannot be eliminated by the attributes of bitmap images. Furthermore, the researchers found that small objects always account for a less proportion of loss in total loss while training \cite{copyenhance}.

In order to improve the detection effect of small objects, previous studies have proposed the following methods: (a) Small objects are copied and randomly placed in other positions of the image to increase the training data samples of small objects during the data augmentation process, which is called replication augmentation \cite{copyenhance}. (b) Images are zoomed and spliced, and some larger objects in the original image are zoomed into small objects. (c) Loss function is designed to pay more attention to small objects by increasing the proportion of small objects' losses \cite{focalloss}.

Due to the problems of scale mismatch and background mismatch in the image processed by using method (a), we only refer to methods (b) and (c) to optimize the training process. The scaling and stitching methods are included in our data augmentation, and the loss function is redesigned, which can effectively improve the detection of small and medium objects and the overall precision of the model.

\section{Approach}

\subsection{Enhanced-Mosaic $\&$ Mixup}

Many real-time object detectors use Mosaic+Mixup strategy in data augmentation during training, which can effectively alleviate the over-fitting situation during training. As shown in Fig. \ref{fig2}(a) and (b), there are two common combination methods, which perform well when a single image in the dataset has relatively sufficient labels. Due to stochastic processes in data argumentation, the data loader might provide images without valid objects while there is response in label space in Fig. \ref{fig2}(a), and the probability of this case increases with the decrease of the label number in each original image.

We design a data augmentation structure in Fig. \ref{fig2}(c). First, we use Mosaic method for several groups of images, and thus the group number can be set according to the richness of the average number of labels in a single picture in the dataset. Then, a last simply processed image is mixed with those Mosaic processed images by Mixup method. In these steps, the original image boundary of our last image is within the boundary of the final output image after transformation. This data augmentation method effectively increases the image richness to alleviate overfitting, and ensures that the output image must contain sufficient effective information.

\begin{figure*}
   \centering
	\includegraphics[width=0.82\linewidth]{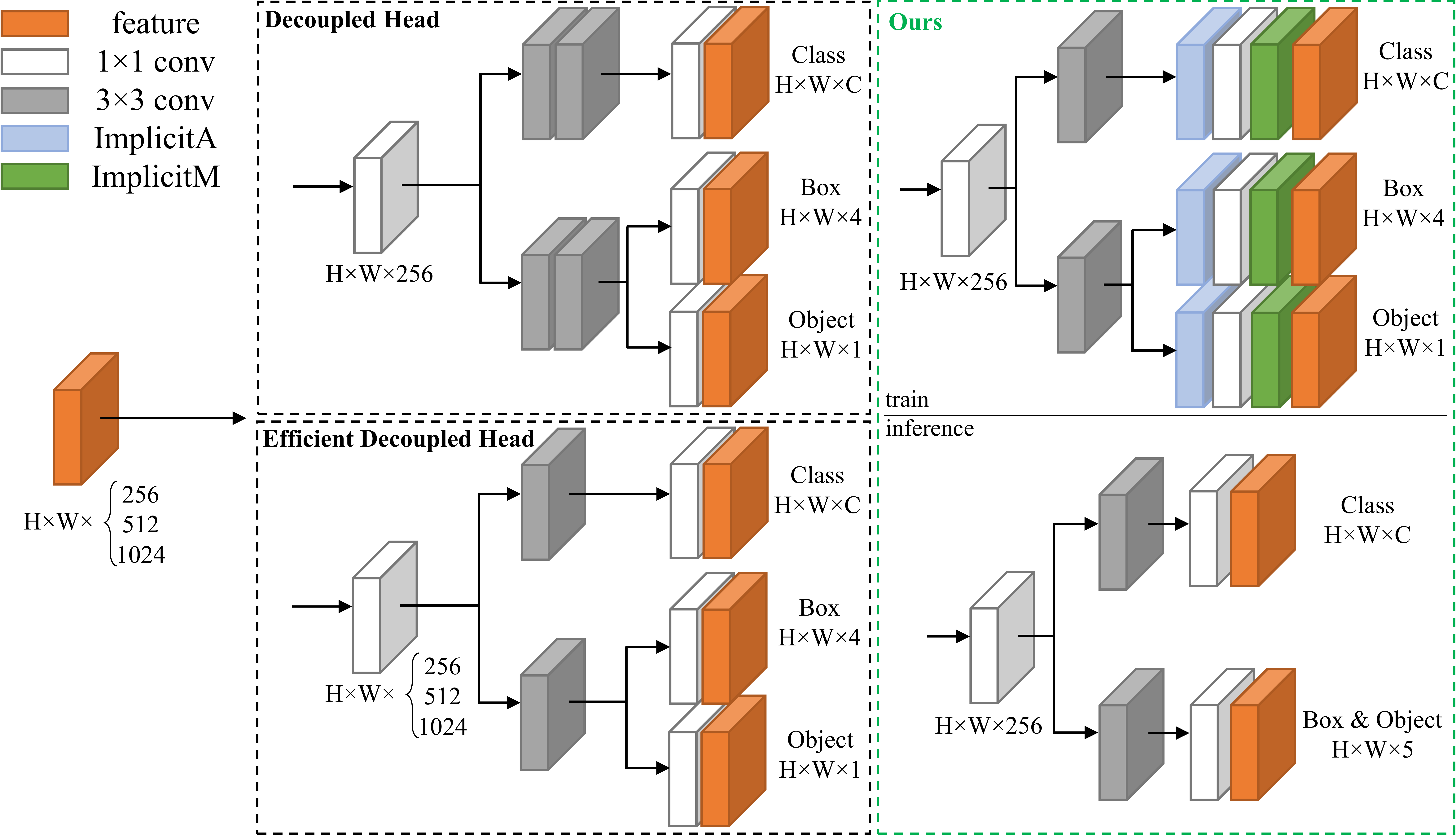}
	\caption{As is shown in this figure, we design a lighter but more efficient decoupled head. With the re-parameterization technique, our model gets a faster inference speed with little precision loss.}
	\label{fig3}
\end{figure*}

\subsection{Lite-Decoupled Head}

Decoupled head in Fig.\ref {fig3} is firstly proposed in FCOS \cite{fcos}, and then used in other anchor-free object detectors, such as YOLOX \cite{yolox}. It is confirmed that using a decoupled structure for the last few network layers can accelerate network convergence and boost the regression performance. 

Since the decoupled head adopts a branch structure that leads to extra inference costs, Efficient Decoupled Head \cite{yolo6} is proposed with a faster inference speed, which reduces the number of the middle 3$\times$3 convolutional layers to only one layer while keeping the same larger number of channels as the input feature map. Nevertheless, in our experiment test, this extra inference cost becomes more apparent with the increase in channels and input size. Thus, we design a lighter decoupled head with fewer channels and convolutional layers. Furthermore, we add implicit representation layers \cite{yolor} to all last convolutional layers for better regression performance. With the method of re-parameterizing, implicit representation layers are integrated into convolutional layers for lower inference costs. The last convolutional layers for box and confidence regression are also merged such that the model can do inference with high parallel computation.

\subsection{Staged Loss Function}

For object detection, the loss function can be generally written as follows
\begin{equation}
  \label{clf}
  L = \alpha L_{cls} + \lambda L_{iou} + \mu L_{obj} + \zeta L_{\Delta}
\end{equation}
where $L_{cls}$, $L_{iou}$, $L_{obj}$ and $L_{\Delta}$ represent classification loss, IOU loss, object loss and regulation loss, and $\alpha, \lambda, \mu, \zeta$ are hyper-parameters. We divide our training process into three stages in our experiments. 

At the first stage, we take one of the most common loss function configurations: gIOU loss for IOU loss, Balanced Cross Entropy loss for classification loss and object loss, and regulation loss setting as zero. The training process steps into the second stage at the last few data-augmentation-enabled epochs. The loss functions of classification loss and object loss are replaced by Hybrid-Random Loss
\begin{equation}
  \label{hrl0}
\begin{aligned}
  {\rm hrl} \left( p, t \right) =& \left[  4 \left(1 - p \right)^2  r + \left( 1 - r \right) \right] t  \log \left( p \right)     \\
								   &+ \left[  12 p^2 r + \left( 1 - r \right) \right] \left( 1 - t \right)  \log \left( 1 - p \right).
\end{aligned}
\end{equation}
where $p$ represents the prediction result, $t$ represents the ground truth and $r$ is a random number between 0 and 1. For all results in one image, we have that
\begin{equation}
  \label{hrl1}
  {\rm HRL} \left( P, T \right) = - \frac{1}{n} \sum_{i=1}^{n} {\rm hrl} \left( P \left( i \right), T \left( i \right) \right)
\end{equation}
which shows a better balance between precision of small objects and total precision. When it comes to the third stage, we close data augmentation and set L1 loss as our regulation loss, and replace gIOU loss by cIOU loss. More details are introduced in the next section.

\section{Experiments}

\newcommand{\tabincell}[2]{\begin{tabular}{@{}#1@{}}#2\end{tabular}}
\begin{table*}[!htb]
  \centering
  \caption{Comparison of different object detectors on COCO 2017-val.}
  \label{coco_map}
  \begin{tabular}{l|c|cccccc|c|c}
   \hhline
   \textbf{Model}  &\textbf{Size} &\textbf{AP$_{val}$} &\textbf{AP$_{50}$} &\textbf{AP$_{75}$} &\textbf{AP$_{S}$} &\textbf{AP$_{M}$} &\textbf{AP$_{L}$} &\textbf{FPS$_{\rm bs=16}$} & \textbf{Params.} \\ \hline
    YOLOv3-ultralytics&640$\times$640& 46.6$\%$                & 66.1$\%$          & 50.4$\%$         & 30.7$\%$     & 51.4$\%$      & 59.1$\%$         & \textbf{37}               & 61.9 M   \\ \hline
    EfficientDet-D6 \cite{edet} & 640$\times$640 & 47.9$\%$    & 67.2$\%$          & -                & -            & -             & -                & 10                        & 51.9 M	\\ \hline
    YOLOv4-CSP      & 640$\times$640 & 47.5$\%$                & 66.2$\%$          & 51.7$\%$         & 28.2$\%$     & 51.2$\%$      & 59.8$\%$         & 25                        & 52.9 M   \\ \hline
    YOLOv5-L        & 640$\times$640 & 48.9$\%$                & 67.6$\%$          & 53.1$\%$         & 31.8$\%$     & 54.5$\%$      & 62.3$\%$         & 30                        & 46.5 M   \\ \hline
    YOLOX-L         & 640$\times$640 & 49.7$\%$                & 68.5$\%$          & 54.5$\%$         & 29.8$\%$     & 54.5$\%$      & 64.4$\%$         & 32                        & 54.2 M   \\ \hline
    EdgeYOLO (\textbf{ours})& 640$\times$640 & \textbf{50.6$\%$} & \textbf{69.8$\%$} & \textbf{54.6$\%$} & \textbf{34.0$\%$} & \textbf{55.1$\%$} & \textbf{65.7$\%$} & 34  & \textbf{40.5 M}  \\ \hline
    \hhline
  \end{tabular}
\end{table*}

\begin{table*}[!htb]
  \centering
  \caption{Comparison of different object detectors on VisDrone2019-DET-val.}
  \label{visdrone_map}
  \begin{tabular}{l|c|cccccc}
   \hhline
   \textbf{Model}  &\textbf{Size} &\textbf{AP$_{val}$} &\textbf{AP$_{50}$} &\textbf{AP$_{75}$} &\textbf{AP$_{S}$} &\textbf{AP$_{M}$} &\textbf{AP$_{L}$} \\ \hline
    YOLOv5-X        & 640$\times$640 & 22.6$\%$                     & 38.6$\%$                    & -                       & -                      & -                      & -                      \\ \hline
    Faster-RCNN + ResNeXt101 \cite{hfpn}& 640$\times$640 & 22.6$\%$ & 40.2$\%$                    & -                       & -                      & -                      & -                      \\ \hline
    Cascade-RCNN + ResNeXt101 \cite{hfpn}& 640$\times$640& 24.4$\%$ & 41.2$\%$                    & -                       & -                      & -                      & -                      \\ \hline
    YOLOX-X         & 640$\times$640 & 25.8$\%$                     & 43.2$\%$                    & \textbf{26.2$\%$}       & 15.9$\%$               & 38.0$\%$               & 52.4$\%$               \\ \hline
    EdgeYOLO (\textbf{ours})& 640$\times$640 & \textbf{26.4$\%$}    & \textbf{44.8$\%$}           & \textbf{26.2$\%$}       & \textbf{16.3$\%$}      & \textbf{38.7$\%$}      & \textbf{53.1$\%$}      \\ \hline
    \hhline
  \end{tabular}
\end{table*}

\subsection{Implementation Details}

\textbf{Dataset.} We test our model's performance on two popular datasets: the common object detection dataset MS COCO2017 and the UAV object detection dataset VisDrone2019-DET, where MS COCO2017 is chosen to be our main training benchmark, and VisDrone2019-DET is chosen specifically for testing the detection performance of small objects.

\textbf{Training.} We deploy the training environment in our workstation with 4 RTX 3090 GPUs. We choose ELAN-Darknet \cite{yolo7} to be our model's backbone and replace some 3$\times$3 convolutional layers with RepConv\cite{repvgg} layers in light models. Our network is trained with the stochastic gradient descent (SGD) optimizer with the maximum learning rate being 0.005 and a batch of 32 images. In particular, our maximum learning rate for each image is fixed to 1/6400, which means the maximum learning rate varies with batch size. We start our training with 5 warm-up epochs with an increasing learning rate ranging from 0 to 0.005. Considering that MS COCO2017 and VisDrone2019-DET have rich labels, the number of mosaic groups is set 2.

\textbf{Inference.} We test inference of each model on edge computing device NVIDIA Jetson AGX Xavier with 512 CUDA cores on MAXN mode, FPS is measured in FP16-precision with TensorRT Version 7.1.3.0.

\textbf{Main training hyper-parameters.}
\begin{itemize}
\item Weight decay: 0.0005
\item Momentum: 0.9
\item Total epochs: 300
\end{itemize}

To view more hyper-parameters, please visit the website: https://github.com/LSH9832/edgeyolo
and all hyper-parameters are in the file ``edgeyolo/train/default.yaml''.

\subsection{Results $\&$ Comparison}

We choose YOLOX-L to be our baseline model, and our detector is compared with some popular object detectors in MS COCO2017-val. All post-process time costs are taken into account when calculating FPS. Additionally, our model is trained in a UAV dataset VisDrone2019-DET \cite{visdrone} which mainly contains small objects, and in this dataset, we compare with some extra-large one-stage models and regular two-stage models which have better performance in MS COCO validation with low inference speed.

Results in MS COCO2017-val are shown in Table~\ref{coco_map}. It can be seen that for device Jetson AGX Xavier, EdgeYOLO is a high-precision real-time detector with less parameters, and its precision even surpasses some no-real-time models. Its total  AP increase mainly comes from detection performance in small objects, which increases 2.2$\%$ AP.

Results in VisDrone2019-DET-val are shown in Table~\ref{visdrone_map}. All of these models are firstly pre-trained on MS COCO2017-train. Even compared with the super models with larger parameters, our model still takes the lead in terms of performance.

\subsection{Ablation Study}

In order to further explore the effects of our approach, we conduct the following experiments on MS COCO2017.

\textbf{Decoupled head.} We compare our model with models using a coupled head and using the original decoupled head. As shown in Table~\ref{headCmp}, by using a lightweight decoupled head, the precision can be significantly improved without losing too much real-time performance.

\begin{table}[h]
  \centering
  \caption{Comparison of different YOLO heads.}
  \label{headCmp}
  \begin{tabular}{l|c|c}
   \hhline
   \textbf{Head}                 &\textbf{AP$_{val}$ ($\%$)} &\textbf{FPS (batch=16)}   \\ \hline
    coupled head                        & 49.8                     & 36   \\ \hline
    original decoupled head             & 50.7                     & 30   \\ \hline
    lite decoupled head (\textbf{ours})  & 50.6                     & 34   \\ \hline
    \hhline
  \end{tabular}
\end{table}

\begin{figure*}
   \centering
	\includegraphics[width=\linewidth]{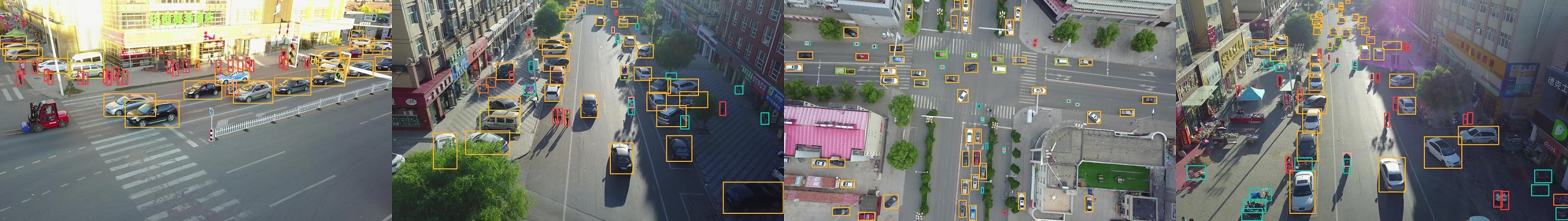}
	\caption{Our representative results in VisDrone2019-DET-val.}
	\label{fig4}
\end{figure*}

\textbf{Segmentation labels (poor effect).} When processing the rotated labels during data augmentation, without segmentation information, we get four coordinate corner points of the original label box after rotation, and draw a box that is not tilted and passes through the four points as the label to be used. This may contain more invalid background information. Therefore, when training our model on MS COCO2017, we try to generate bounding boxes by using segmentation labels, so that the labels after image rotation still maintain high accuracy. When the data augmentation is enabled and the loss enters a stable decline phase, using segmentation labels can bring a significant increase by 2$\%$ - 3$\%$  AP. Since the data augmentation is set disabled at the last stage of training, all labels become more accurate. Moreover, even if the segmentation labels are not used, the final accuracy decreases only by about 0.04$\%$  AP.

\textbf{Loss function.} We have repeatedly compared and tested the methods with various loss functions. As shown in Table~\ref{lossAblation}, focal loss \cite{focalloss}, which is designed to improve the sample imbalance problem, plays an opposite role in our model-training. Another interesting phenomenon is that, when HR loss is used, compared with using cIOU, using gIOU causes a decrease in precision of large objects by 0.7$\%$ AP despite the fact that it can increase the precision of small objects by 0.3$\%$ AP. To sum up, a better precision can be obtained by using HR loss and cIOU loss in later training stages.

\begin{table}[h]
	\centering
	\caption{Comparison of using different losses.}
	\label{lossAblation}
	\begin{tabular}{l|cccc}
		\hhline
		\textbf{Loss}   &\textbf{AP$_{val}$ ($\%$)} &\textbf{AP$_{S}$} &\textbf{AP$_{M}$} &\textbf{AP$_{L}$}   \\ \hline
		BCE + gIOU (baseline) &50.1                      &33.6              &54.7              &65.4                \\ \hline
		Focal + gIOU         &49.4                      &33.3              &53.9              &63.3                \\ \hline
		HR + gIOU            &50.4                      &\textbf{34.3}     &\textbf{55.1}     &65.0                \\ \hline
		BCE + cIOU           &50.2                      &33.8              &54.8              &63.3                \\ \hline
		Focal + cIOU         &49.2                      &33.2              &53.7              &63.5                \\ \hline
		HR + cIOU            &\textbf{50.6}             &34.0              &\textbf{55.1}     &\textbf{65.7}       \\ \hline
		\hhline
	\end{tabular}
\end{table}

\subsection{Tricks for Edge Computing Devices}

\textbf{Input size adaptation.} In practice, the input source of the object detection algorithm on edge computing devices is often a video stream with a fixed size and aspect ratio. At present, the commonly used video stream aspect ratios are 4:3 and 16:9. As in Table~\ref{inputSize}, when it is applied to practice, the network input size is usually set to 640$\times$480 and 640$\times$384 to reduce the amount of model computation, which can significantly improve the inference speed without losing accuracy.

\begin{table}[h]
  \centering
  \caption{Comparison of different network input sizes.}
  \label{inputSize}
  \begin{tabular}{c|c|c}
   \hhline
   \textbf{Video Aspect Ratio} &\textbf{Input Size}       &\textbf{FPS (batch=16)}   \\ \hline
    1:1                        & 640$\times$640                     & 34   \\ \hline
    4:3                        & 640$\times$480                     & 43   \\ \hline
    16:9                       & 640$\times$384                     & 50   \\ \hline
    \hhline
  \end{tabular}
\end{table}

\textbf{Multi-process ${\rm \&}$ multi-thread computing architecture.} As a whole detection process containing pre-process, model input and post-process, these three parts can be split in actual deployment and allocated to multiple processes and threads for calculation. In our test, using a split architecture can achieve about 8$\%$-14$\%$ FPS increase.

\begin{figure}
   \centering
	\includegraphics[width=\linewidth]{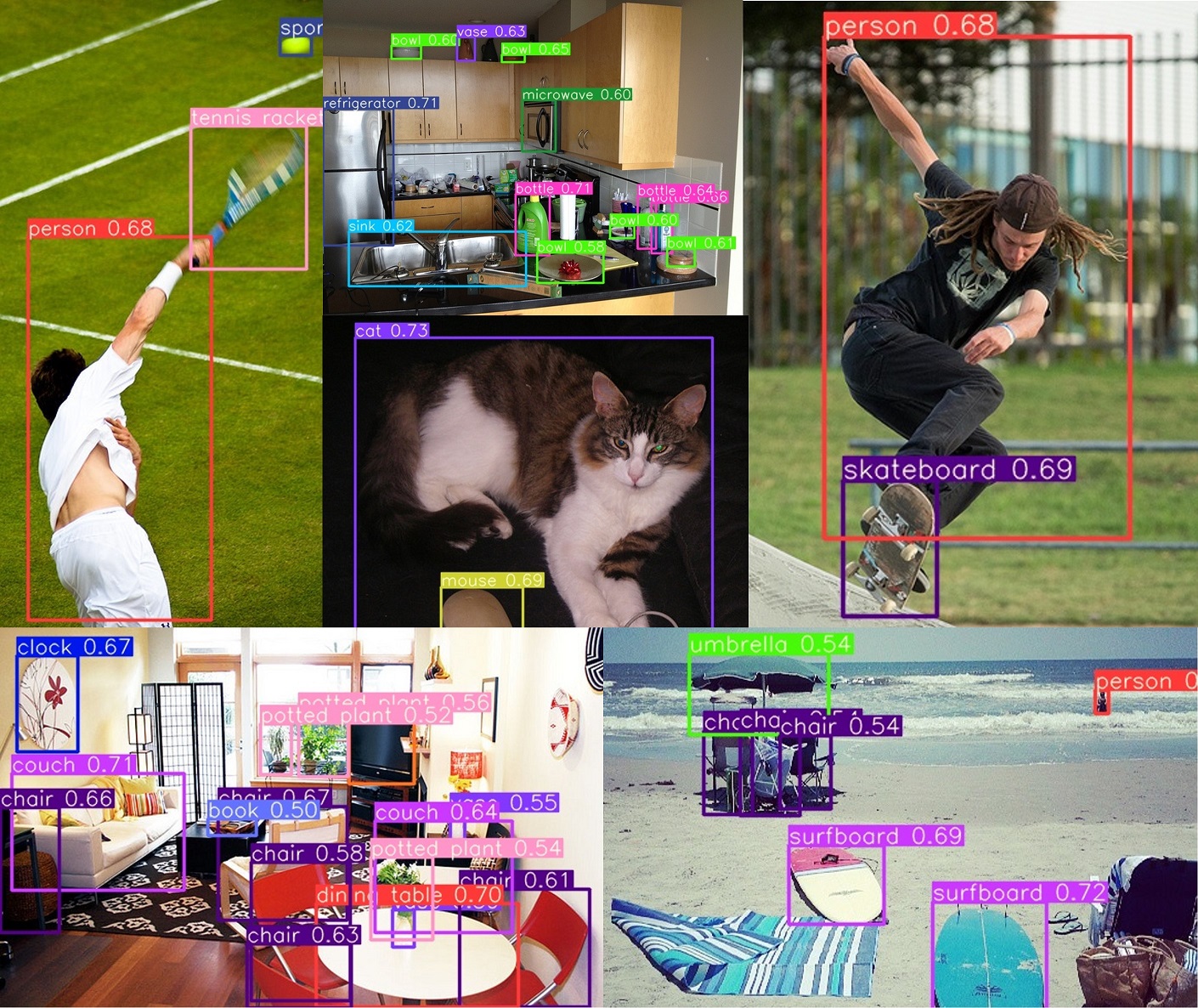}
	\caption{Our representative results on MS COCO2017-val.}
	\label{fig5}
\end{figure}

\section{Conclusion}

We have proposed an edge-real-time and anchor-free one-stage detector EdgeYOLO, some representative results of which are shown in Fig. \ref{fig4} and Fig. \ref{fig5}. As shown in the experiment, EdgeYOLO can run on edge devices in real time with high accuracy, and its ability to detect small objects has been further improved. Since EdgeYOLO uses an anchor free structure, the design complexity and computational complexity are reduced, and the deployment on edge devices is more friendly. Moreover, we believe that the framework can be extended to other pixel level recognition tasks such as instance segmentation. In future work, we will further improve the detection accuracy of the framework for small objects, and carry out explorations for efficient optimizations.


\end{document}